# Prostate Cancer Diagnosis using Deep Learning with 3D Multiparametric MRI

Saifeng Liu*[a], Huaixiu Zheng*[b], Yesu Feng[c], Wei Li[d]

[a] The MRI Institute for Biomedical Research, Waterloo, ON, Canada; [b] Uber Technologies Inc., San Francisco, CA, USA; [c] LinkedIn, San Francisco, CA, USA; [d] USAA, San Antonio, TX, USA

## ABSTRACT

A novel deep learning architecture (XmasNet) based on convolutional neural networks was developed for the classification of prostate cancer lesions, using the 3D multiparametric MRI data provided by the PROSTATEx challenge. End-to-end training was performed for XmasNet, with data augmentation done through 3D rotation and slicing, in order to incorporate the 3D information of the lesion. XmasNet outperformed traditional machine learning models based on engineered features, for both train and test data. For the test data, XmasNet outperformed 69 methods from 33 participating groups and achieved the second highest AUC (0.84) in the PROSTATEx challenge. This study shows the great potential of deep learning for cancer imaging.

## 1. INTRODUCTION

Prostate cancer is the second most common type of cancer in men[1]. It is estimated that there will be 161,360 new cases and 26,730 deaths caused by prostate cancer in US in 2017[2]. The screening of prostate cancer using Prostate Specific Antigen (PSA) test and Digital Rectal Examination (DRE) has very limited diagnostic accuracy and may lead to over-diagnosis[3]. Although the prostate cancer is usually not the direct cause of death, the differentiation of clinically significant prostate cancer lesions from those low-grade lesions is critical[1]. The use of multiparametric MRI (mpMRI) with Prostate Imaging Reporting And Data System (PI-RADS) has increased the sensitivity in prostate cancer detection, but the specificity is still poor which may lead to unnecessary biopsies[4,5]. The accuracy of mpMRI with PI-RADS is also largely dependent on the experience of the radiologist[5]. In the last few years, several machine learning algorithms have been proposed[6,7], based on engineered features. However, feature engineering usually requires a significant amount of domain expertise and hence the accuracies of these models are highly variable depending on the quality of the engineered features.

The advent of deep convolutional neural network (CNN) techniques has led to great success in natural image classification[8,9]. For medical imaging, however, the effectiveness of deep learning is hampered by two major challenges: the heterogeneous raw data and the relatively small sample size[10]. In this paper, these two challenges were conquered through proper data preprocessing and data augmentation. We develop a new deep CNN architecture (XmasNet) and performed an end-to-end training of XmasNet on 3D multiparametric MRI images for prostate cancer diagnosis. The rest of the paper is organized as follows. Section 2 describes the details of data preprocessing and data augmentation. Section 3 presents the architecture of XmasNet and the end-to-end training. Section 4 shows feature engineering and the training of conventional machine learning model. The results and discussions are provided in Section 5, while conclusions and future directions are given in Section 6.

## 2. DATA PREPROCESSING AND DATA AUGMENTATION

In this study, there were 341 cases, including 204 training cases, in which the labels of the lesions (clinically significant or non-significant) were given. For each case, there were Diffusion Weighted Images (DWI), Apparent Diffusion Coefficient (ADC) maps, Ktrans, as well as T2 Weighted Images (T2WI). At least one lesion was present in each case, and the locations of the lesions were provided. There were three major steps in data preprocessing: 1. *Standardization*.

---

The images acquired with different parameters were interpolated to 1mm isotropic resolution, using linear interpolation. For each case, different types of images were co-registered, using the transverse T2WI as the reference. 2. *Refining lesion center*. The lesion region was obtained through region growing and morphological operations, using DWI. Then the center of the lesion was refined as the centroid of the obtained region. 3. *Train/validation sample preparation*. Five cases were removed from the training data, due to their relatively low image quality. The remaining 199 cases were split into train and validation sets using a stratified splitting. Specifically, there were 169 cases (274 lesions) for training and 30 cases (43 lesions) for validation. Four different types of inputs were generated, using different combinations of DWI (D), ADC (A), Ktrans (K), and transverse T2WI (T) as the RGB channels: DAK, DAT, AKT, DKT (Figure 1). For each type of input, at least one deep learning model was trained.

Data augmentation was done through 3D rotation and slicing. Particularly, we sliced the lesion at 7 different orientations. For each slice, in-plane rotation, random shearing and translation of the lesion to ±1 pixel were also performed. In the end, 207144 train samples were prepared, with each sample corresponding to a 32×32 region of interest (ROI) surrounding the lesion center. When training using Caffe[11], random mirroring was also performed on the fly. The same augmentation procedure was also applied to validation and test sets. For each lesion in the test set, the final prediction was an average of the predictions made from different orientations. This multi-view technique reformulates the 3D problem into a 2D problem, and enables the incorporation of 3D information in the 2D inputs.

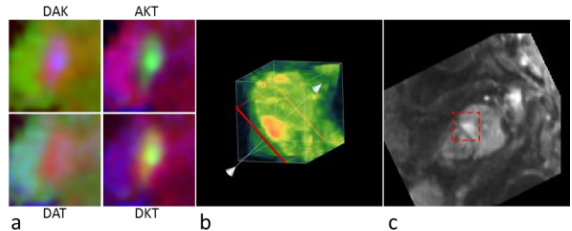

Figure 1. a. Examples of the four types of input images for XmasNet. See main text for definitions of the input types. b. Illustration of data augmentation through 3D slicing. c. Illustration of data augmentation through in-plane rotation. The red dashed box in c shows the cropped 32×32 region of interest (ROI) centered at the lesion.

### 3. DEEP LEARNING MODEL: XMASNET

A new deep learning architecture, named XmasNet (Figure 2), was built for this study. XmasNet was inspired by the VGG net[9]. The input and output sizes of each layer are listed in Table 1. The end-to-end training of XmasNet was carried out on an AWS instance.

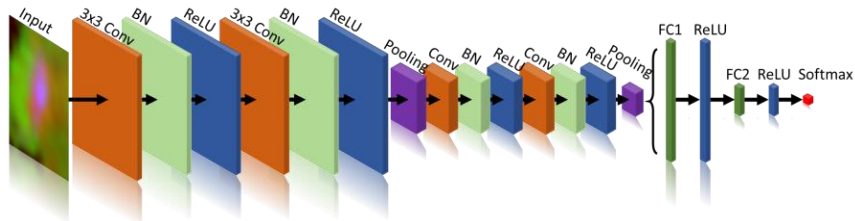

Figure 2. The architecture of the XmasNet. Conv: convolutional layer; BN: batch normalization layer; ReLU: rectified linear unit; Pooling: max pooling layer; FC: fully connected layer.

The process of training the XmasNet with DAK input is illustrated in Figure 3.a. For training this net, a learning rate of $2\times10^{-6}$ was used, with Adam solver and weight decay being $10^{-4}$. The best validation AUC (0.92) occurred at the $1610^{th}$ step, and thus we stopped the training at that step. Similarly, 20 XmasNet models with other types of input were trained. For the lesions in the test set, the predictions were obtained as the weighted average of the predictions from individual XmasNet models. In order to determine the weights of different nets, a greedy bagging algorithm[12] was used. Specifically, in each iteration, the model which maximizes the validation AUC of the ensemble was selected from the 20 models and added to the ensemble. The iteration was stopped when the validation AUC converged, and the final weight of a model was proportional to the number of times that model being selected. As shown in Figure 3.b, 5 models with relatively higher AUC and 2 models with relatively lower AUC were included in the model ensemble. The inclusion of the two weaker models is because of the complementary information they provided to the other models. Then the final prediction for each lesion was calculated as the average of the predictions made from different orientations.

Table 1. The patch size, stride and output size of each layer in XmasNet.

| Layers | Conv1 | Conv2 | Max pooling1 | Conv3 | Conv4 | Max pooling2 | FC1 | FC2 | Softmax |
|---|---|---|---|---|---|---|---|---|---|
| **Patch size/stride** | 3×3 / 1 | 3×3 / 1 | 2×2 / 2 | 3×3 / 1 | 3×3 / 1 | 2×2 / 2 | | | |
| **Output size** | 32×32 ×32 | 32×32 ×32 | 16×16 ×32 | 16×16 ×64 | 16×16 ×64 | 8×8 ×64 | 1024×1 | 256×1 | 2×1 |

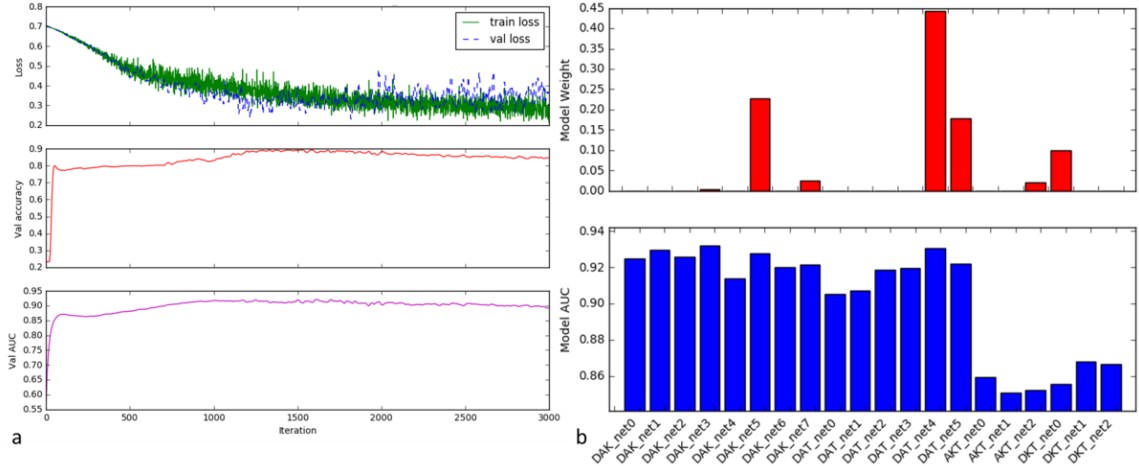

Figure 3. a. Training the XmasNet with DAK input. The best validation AUC (0.92) occurred at the 1610$^{th}$ step. b. The XmasNets and their weights in the model ensemble. Seven XmasNets with non-zero weights were included in the final ensemble, including five models with relatively higher AUC and two models with relatively lower AUC.

## 4. FEATURE ENGINEERING AND DECISION TREE MODEL

We extracted a total of 87 features for each lesion, including means and standard deviations of the intensities, as well as texture features such as energy, contrast etc., obtained from a 3D Haralick texture analysis[13]. 140 decision tree models were trained using a gradient boosting algorithm, XGBoost[14], with backward feature selection and hyper-parameter tuning performed independently for each model through 5-fold cross validation. Finally, the top 20 models with high cross-validation AUC were included in the model ensemble obtained through a similar greedy bagging procedure[12] as XmasNet models. We observed that the frequently appearing features were usually related to functional information such as Ktrans and ADC, and were mostly texture features. Specifically, four features appearing most frequently are the correlation, variance and the minimum intensity of the lesion region on Ktrans image, as well as the minimum intensity of the lesion region on ADC map. These features were found to be selected by all the top 20 models.

## 5. COMPARISON OF DEEP LEARNNG AND XGBOOST MODELS

Figure 4.a shows the performance of the XmasNet and the XGBoost models on the validation data. The ROC curve for XmasNet was obtained from the validation set with data augmentation, while the ROC curve for XGBoost model was obtained from the whole train data set with 5-fold cross validation. For both train and test data, deep learning model outperformed conventional machine learning model based on engineered features by a significant margin. In the train set, the AUCs were 0.95 and 0.89, for XmasNet and XGBoost model, respectively; in the test set, the AUC was 0.84 and 0.80, for the XmasNet and XGBoost model, respectively. The drop in AUCs from train to test data is most likely due to overfitting caused by the relatively small sample size of the train data. Yet the AUC (0.84) of the XmasNet on the test data is higher than the AUC (0.83) obtained from mpMRI with PIRADS V2 reported in a previous study[5]. In that study, a sensitivity of 0.77 and a specificity of 0.81 were obtained. Using XmasNet on the validation data, both the sensitivity and the specificity were improved to 0.89. On the other hand, using the XGBoost model, when the sensitivity was 0.87,

which was higher than the sensitivity of the mpMRI model, the specificity was 0.77, which was slightly lower than that of the mpMRI model (Figure 4.b).

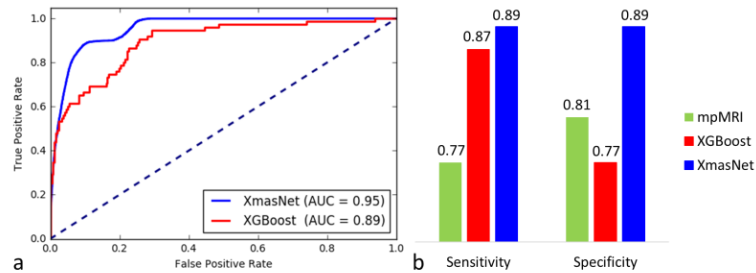

Figure 4. a. The ROC curves of XmasNet and XGBoost model on the validation data. b. Comparison of the sensitivities and specificities of different models.

## 6. CONCLUSIONS AND FUTURE DIRECTIONS

In this study, a new deep learning architecture, XmasNet, was built for prostate cancer diagnosis. The data augmentation was done through 3D rotation and slicing, in order to avoid over-fitting and to capture the 3D information. We performed end-to-end training for XmasNet. It outperformed 69 methods from 33 participating groups, and had the second highest AUC in 2017 PROSTATEx challenge. The results also showed that high sensitivity and specificity can be achieved using the proposed model. XmasNet can be further improved by combining hand-crafted features and by using more sophisticated architectures. It was observed that the deep learning model outperformed the conventional model based on feature engineering in both train and test sets. This study demonstrates the great potential of deep learning for cancer imaging.